\documentclass[lettersize,journal]{IEEEtran}
\usepackage{amsmath,amsfonts}
\usepackage{algorithmic}
\usepackage{algorithm}
\usepackage{array}
\usepackage[caption=false,font=normalsize,labelfont=sf,textfont=sf]{subfig}
\usepackage{textcomp}
\usepackage{stfloats}
\usepackage{url}
\usepackage{hyperref}
\usepackage{verbatim}
\usepackage{graphicx}
\usepackage{cite}
\usepackage{cleveref}
\crefformat{equation}{(#2#1#3)}
\crefformat{equation}{Eq.~(#2#1#3)}
\crefformat{figure}{Fig.~#2#1#3}
\crefformat{algorithm}{Alg.~#2#1#3}
\crefformat{table}{Tab.~#2#1#3}
\crefformat{section}{Sec.~#2#1#3}
\crefrangeformat{equation}{(#3#1#4) to~(#5#2#6)}
\crefmultiformat{equation}{(#2#1#3)}%
{ and~(#2#1#3)}{, (#2#1#3)}{ and~(#2#1#3)}

\usepackage{multirow}
\usepackage[normalem]{ulem}
\usepackage[flushleft]{threeparttable}
\usepackage{tabularx}

\usepackage{xcolor}

\begin{document}

\title{Talk2DM: Enabling Natural Language Querying and Commonsense Reasoning for Vehicle-Road-Cloud Integrated Dynamic Maps with \\Large Language Models}

\author{
Lu Tao, 
Jinxuan Luo,
Yousuke Watanabe, 
Zhengshu Zhou,
Yuhuan Lu,
Shen Ying,
Pan Zhang,
Fei Zhao,\\
Hiroaki Takada

\thanks{Lu Tao and Shen Ying are with the School of Resource and Environmental Sciences, Wuhan University, Wuhan, China.}
\thanks{Jinxuan Luo and Fei Zhao are with the School of Earth Sciences, Yunnan University, Kunming, China.}
\thanks{Zhengshu Zhou is with the College of Artificial Intelligence, Tianjin University of Science \& Technology, Tianjin, China.}
\thanks{Yuhuan Lu is with the Department of Computer and Information Engineering, Khalifa University, Abu Dhabi, United Arab Emirates.}
\thanks{Pan Zhang is with NVIDIA, Shanghai, Chian.}
\thanks{Yousuke Watanabe and Hiroaki Takada are with the Institutes of Innovation for Future Society, Nagoya University, Nagoya, Japan.}

}


\maketitle

\begin{abstract}
Dynamic maps (DM) serve as the fundamental information infrastructure for vehicle-road-cloud (VRC) cooperative autonomous driving in China and Japan. By providing comprehensive traffic scene representations, DM overcome the limitations of standalone autonomous driving systems (ADS), such as physical occlusions. Although DM-enhanced ADS have been successfully deployed in real-world applications in Japan, existing DM systems still lack a natural-language-supported (NLS) human interface, which could substantially enhance human-DM interaction. To address this gap, this paper introduces VRCsim, a VRC cooperative perception (CP) simulation framework designed to generate streaming VRC-CP data. Based on VRCsim, we construct a question-answering data set, VRC-QA, focused on spatial querying and reasoning in mixed-traffic scenes. Building upon VRCsim and VRC-QA, we further propose Talk2DM, a plug-and-play module that extends VRC-DM systems with NLS querying and commonsense reasoning capabilities. Talk2DM is built upon a novel chain-of-prompt (CoP) mechanism that progressively integrates human-defined rules with the commonsense knowledge of large language models (LLMs). Experiments on VRC-QA show that Talk2DM can seamlessly switch across different LLMs while maintaining high NLS query accuracy, demonstrating strong generalization capability. Although larger models tend to achieve higher accuracy, they incur significant efficiency degradation. Our results reveal that Talk2DM, powered by Qwen3:8B, Gemma3:27B, and GPT-oss models, achieves over 93\% NLS query accuracy with an average response time of only 2$\sim$5 seconds, indicating strong practical potential.
\end{abstract}


\section{Introduction}
\begin{figure}[htpb]
	\centering
	\includegraphics[scale=0.32]{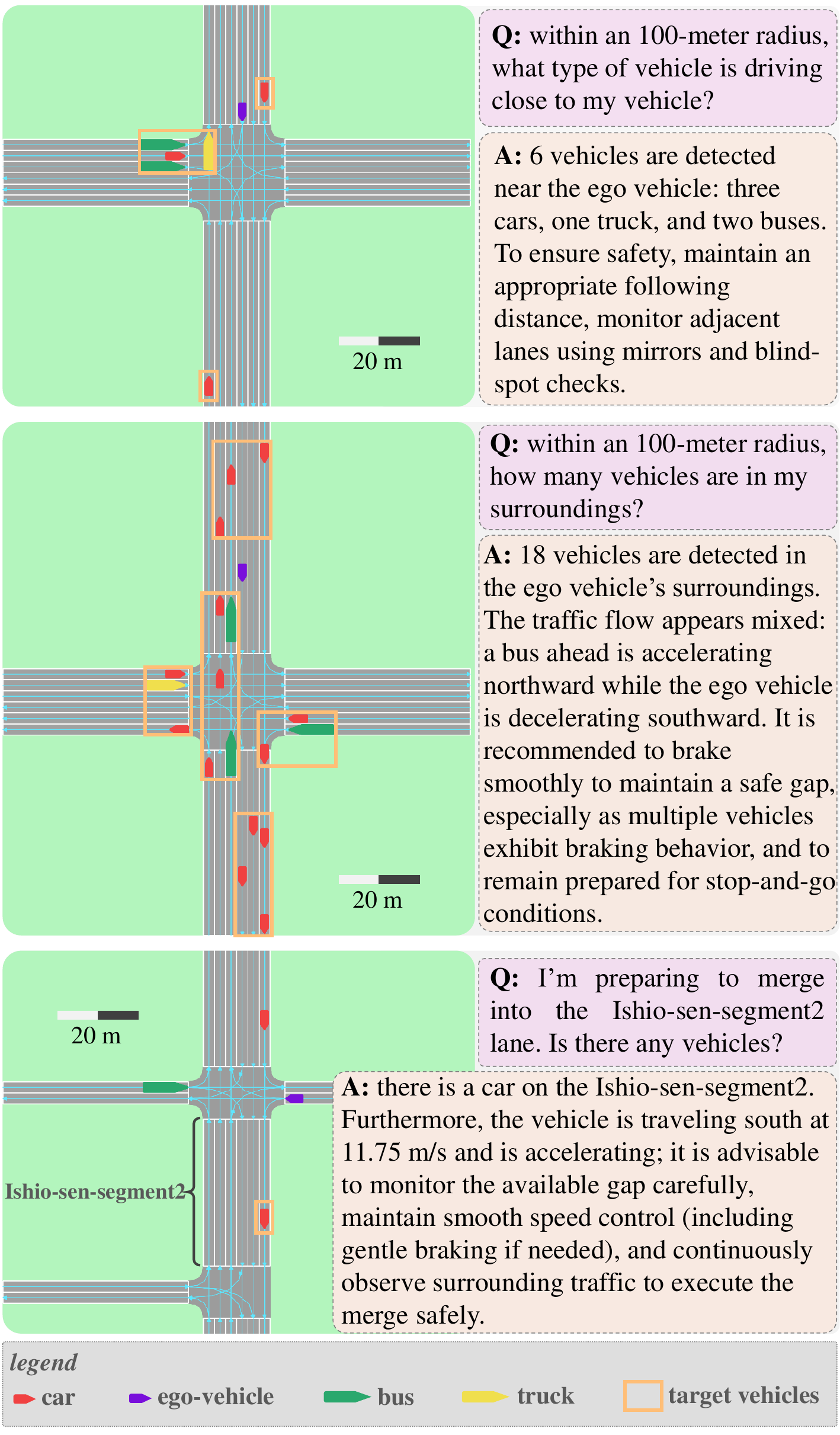}
	\caption{Talk2DM--A human interface of VRC-DM framework, enabling over-the-horizon perception, natural-language-supported querying, and commonsense reasoning. A demonstration video of Talk2DM can be viewed at \href{https://youtu.be/mg3UsLoHz2Q}{Youtube} or \href{https://www.bilibili.com/video/BV1Crc4zAE5T}{Bilibili}}.
	\label{fig-example}
\end{figure}

\IEEEPARstart{T}{he} emergence of Large Language Models (LLMs) has significantly sparked a paradigm shift in the field of Autonomous Driving (AD), involving perception, prediction and planning tasks \cite{choudhary2024talk2bev,ding2024holistic, peng2025lc, mao2023gpt}. Compared with conventional AD systems (ADS), LLMs-enhanced ADS offer greater interoperability, interpretability, and flexibility between humans and vehicles, thus gaining more and more attentions in both academia and industry. However, most studies on LLM-enhanced AD (LLM-AD) primarily focus on a single Autonomous Vehicle (AV), owing to the lack of Question-Answer (QA)-annotated cooperative AD datasets. Recently, NVIDIA has released a Vehicle-to-Vehicle (V2V) QA dataset , V2V-QA \cite{chiu2025v2v}, bridging the gap between LLM-AD and cooperative AD. However, their study only focus on two-vehicle V2V collaboration in grounding and planing tasks when AVs' Line-of-Sight (LoS) and Field-of-View (FoV) are blocked. In China, the Vehicle-Road-Cloud Integration System (VRCIS) is regarded as the next AD stage that comprehensively enhances vehicle performance, traffic safety and efficiency \cite{xu2022status,gao2024vehicle}. Unfortunately, to date, the study on LLM-enhanced VRCIS remains unexplored due to the absence of an open QA-annotated VRCIS dataset.

Dynamic maps (DM) \cite{watanabe2020dynamicmap} serve as the fundamental information infrastructure for cooperative ADS in Japan, integrating cloud, edge and embedded systems. God's Eye View (GEV) \cite{tao2024enhancing} is a DM-embedded data representation that extends the widely-known Bird's Eye View (BEV) concept in both spatial and temporal domains through the Cooperative Perception (CP) and Cooperative Prediction (CPd) modules within a Vehicle-Road-Cloud (VRC) integrated Dynamic Map (VRC-DM) framework. GEV represents structured cooperative perception and prediction data on VRC-DM platforms, in contrast to conventional BEV, which rasterizes the local environment using onboard sensor observations from a single AV. By leveraging cooperative information, GEV overcomes inherent BEV limitations, such as Line-of-Sight (LoS), Field-of-View (FoV) and Range-of-Time (RoT), and demonstrates greater potential for comprehensive traffic scene representation.
DM is built upon relational models, enabling standardized information query and  data operation using SQL-like languages with high efficiency, e.g., prediction\cite{tao2024enhancing}. However, the standardization weaken its flexibility. Recent studies on LLM-AD \cite{choudhary2024talk2bev, ding2024holistic, chiu2025v2v} highlight the strong commonsense reasoning, out-of-distribution generalization, knowledge retrieval abilities and broad applicability of pre-trained LLMs in AD. The comprehension and reasoning capabilities, along with the feature of natural language input, greatly benefit their flexibility and diversity in information retrieval. These strengths, however, are largely absent in SQL-based DM platforms.

Current LLM-AD studies can be viewed as an adaption of the classical Visual Question Answering (VQA) task \cite{antol2015vqa} to AD domain, both entailing modality conversion from image to text. There are two typical paradigms: modular and integrated methods. The modular methods \cite{choudhary2024talk2bev, ding2024holistic, shao2023prompting, ozdemir2024enhancing, yang2022empirical, guo2023from} add a plug-and-play pre-processing module before the frozen LLMs, making them model-agnostic. The pre-processing module can be a pre-trained caption model \cite{yang2022empirical, guo2023from, ozdemir2024enhancing, choudhary2024talk2bev}, an off-the-shelf VQA model \cite{shao2023prompting}, or a trainable injection module \cite{ding2024holistic}, all of which convert images into textual input for the LLMs. This configuration brings two benefits \cite{guo2023from}: first, it reduces the deployment cost and simplify the deployment; second, upgrading the underlying LLMs is straightforward. The integrated methods finetune a vision encoder jointly with the LLMs to align visual and linguistic representations \cite{xu2024drivegpt4, chiu2025v2v, wu2025language, qian2024nuscenes, chen2024driving}. These methods finetune LLMs' parameters to tailor integrated models for various AD tasks, potentially harming the generalization and commonsense reasoning ability of the LLMs. In addition, funetuning LLMs incur prohibitive computational and data cost, which is unaffordable for most researchers.

To bridge the aforementioned gaps, we propose Talk2DM, a plug-and-play module designed to enable natural-language querying and commonsense reasoning on VRC-DM platforms. Specifically, we first introduce a VRC simulation framework (VRCsim) that supports structured CP data generation in a VRC collaborative manner, closely mimicking the real-world deployed VRC-DM systems for Level4 AD in mixed-traffic environments \cite{tao2024enhancing}. Second, we construct a VRC Question Answering dataset (VRC-QA) comprising over 10K mixed-traffic scenes and 100K QA pairs. Third, we develop Talk2DM as a baseline LLM plugin for VRC-DM platforms, enhancing flexibility and diversity in information retrieval while endowing the system with commonsense reasoning capabilities. Finally, we conduct comprehensive evaluations of Talk2DM on the VRC-QA dataset and establish a VRC-QA benchmark to facilitate future research on LLMs for VRC-cooperative AD. Our contributions are summarized as follows:
\begin{itemize}
	\item \textit{\textbf{VRCsim}} is devised for simulating Vehicle-Road-Cloud Cooperative Perception (VRC-CP). It generates data streams under collaborative scenarios involving Road Side Units (RSUs), AVs and cloud nodes, where perception data from RSUs and AVs are aggregated in the cloud. The cloud node subsequently reconstructs digital scenes and renders them into textual representations, enabling downstream information retrieval and commonsense reasoning through LLMs. To the best of our knowledge, VRCsim is the first simulation framework specifically designed for VRCIS.   
	
	\item \textit{\textbf{VRC-QA}} is created to support the development and evaluation of LLM-enhanced VRC-DM, with a particular focus on spatial querying and reasoning in mixed-traffic scenes. VRC-QA comprises over 10K VRC-CP scenes paired with more than 100K QA pairs and has been made publicly available\footnote{\url{https://github.com/LuTaonuwhu/VRC-QA.git}}. 
	
	\item  \textit{\textbf{Talk2DM}}, as demonstrated in \cref{fig-example}, is proposed to enhance the information retrieval flexibility of VRC-DM and to equip it with commonsense reasoning capabilities. Talk2DM enables Over-The-Horizon (OTH), Natural-Language-Supported (NLS) querying and commonsense reasoning. Moreover, it operates in a plug-and-play manner and leverages off-the-shelf pre-trained LLMs, allowing for painless generalization as well as low-cost development and deployment. 
\end{itemize}

The remainder of this paper is organized as follows: section 2 reviews related work; section 3 introduces the system architecture; section 4 presents the experiments and section 5 concludes this paper.
 
\section{Related Work}
\subsection{Large Language Model Enhanced Autonomous Driving}
Integrating LLMs and Vision-Language Models (VLMs) with ADS has attracted widespread attention due to their remarkable potential in scene understanding, logical reasoning and response generating \cite{zhou2024vision,cui2024survey}. Pioneering works, such as Cityscapes-Ref\cite{vasudevan2018object} and Talk2Car \cite{deruyttere2019talk2car}, attempt to incorporate natural language into object detection tasks in AD. However, both Cityscapes-Ref and Talk2Car limit each expression to refer to a individual object in each image. To overcome this limitation, Refer-KITTI \cite{wu2023referring} is developed, where each prompt can refer to multiple objects. Recently, NuScene-QA \cite{qian2024nuscenes} has made a significant stride, where VQA is adopted to propel referring expression understanding into high-level driving scenario understanding. Concurrently, Talk2BEV \cite{choudhary2024talk2bev} is proposed to augment BEV maps with natural language, enabling general-propose visuolinguistic reasoning for AD scenarios. In more fine-grained 3D object tracking tasks, NuPrompt is constructed and an end-to-end 3D tracker, PromptTrack, is proposed to match referent objects following human instructions \cite{wu2025language}. In \cite{ding2024holistic}, a holistic language-driving dataset, NuInstruct, is curated; and BEV-InMLLM is proposed to integrate instruction-aware BEV features with pre-trained MLLMs. In \cite{xu2024drivegpt4}, DriveGPT4 is presented for interpretable end-to-end AD; it can process multimodal input data and generate text responses as well as low-level control signals, such as speed and turning angle. 
These studies focus on merging visual and textual modalities for LLM adaptation in individual AV scenarios. 

Different from them, \cite{chen2024driving} proposes an object-level multimodal LLM architecture to fuse object-level numeric data with LLMs, bypassing the visual features extraction and alignment. V2V-LLM \cite{chiu2025v2v} is a pioneering study that dedicates to V2V cooperative AD utilizing LLMs, in which scene-level feature maps and object-level feature vectors obtained by multiple connected AVs (CAVs) are fused and fed to central LLMs to answer grounding and planning related questions for CAVs.

This paper is closely related to Talk2BEV \cite{choudhary2024talk2bev}, V2V-LLM \cite{chiu2025v2v} and LLM-Driver \cite{chen2024driving}. Following these insightful studies, we mainly focus on OTH and NLS querying as well as commonsense reasoning in a VRC cooperative framework, an uncharted aspect in the literature. A comparison with related work is presented in \cref{tab-compare}.

\begin{table*}[htpb]
\centering
\begin{threeparttable}
	\caption{Comparison with related studies.}
	\label{tab-compare}
	\begin{tabular}{cccccccccc}
		\hline \hline
		\multicolumn{1}{c}{\multirow{2}{*}{\textbf{Model}}} & \multicolumn{1}{c}{\multirow{2}{*}{\textbf{Publication}}} & \multicolumn{1}{c}{\multirow{2}{*}{\textbf{Domain}}} & \multicolumn{4}{c}{\textbf{Input modality}}                                                                                                                                                                                                                                & \multicolumn{2}{c}{\textbf{Methodogy}}                                                                           & \multicolumn{1}{c}{\multirow{2}{*}{\textbf{\begin{tabular}[c]{@{}c@{}}Backbone \\ LLMs / Models\end{tabular}}}} \\ \cline{4-9}
		\multicolumn{1}{c}{}                                & \multicolumn{1}{c}{}                                 & \multicolumn{1}{c}{}                                 & \multicolumn{1}{c}{\begin{tabular}[c]{@{}c@{}}image /\\ video\end{tabular}} & \multicolumn{1}{c}{\begin{tabular}[c]{@{}c@{}}point \\ cloud\end{tabular}} & \multicolumn{1}{c}{text} & \multicolumn{1}{c}{\begin{tabular}[c]{@{}c@{}}structured \\ object data\end{tabular}} & \multicolumn{1}{c}{\begin{tabular}[c]{@{}c@{}}training /\\ fine-turning\end{tabular}} & \multicolumn{1}{c}{prompting} & \multicolumn{1}{c}{}                                                                                          \\ \hline
		PICa\cite{yang2022empirical} & AAAI 2022 & VQA & $\surd$ & & $\surd$ & & & $\surd$  & GPT-3\\ 
		Prophet\cite{shao2023prompting}   & CVPR 2023  & VQA  & $\surd$ & & $\surd$  &  &    & $\surd$  & GPT-3  \\
		Img2LLM\cite{guo2023from} & CVPR 2023 & VQA  & $\surd$  & & $\surd$ &   &  & $\surd$  & \begin{tabular}[c]{@{}c@{}}OPT, GPT-J,\\ GPT-Neo, BLOOM\end{tabular}\\ 
		DriveGPT4\cite{xu2024drivegpt4} & IEEE RAL 2024 & AV(C) & $\surd$ &   & $\surd$ &  & $\surd$   & & LLaMA2 \\
		BEV-InMLLM\cite{ding2024holistic}& CVPR 2024 & AV(3P) & $\surd$ &  & $\surd$ & & $\surd$ & & \begin{tabular}[c]{@{}c@{}}BLIP-2,MiniGPT-4, \\ Video-Llama\end{tabular} \\ 
		Talk2BEV\cite{choudhary2024talk2bev} & ICRA 2024 & AV(P1P3)  & $\surd$ & $\surd$ & $\surd$ &   &    & $\surd$ & \begin{tabular}[c]{@{}c@{}}BLIP-2, MiniGPT-4\\ InstructBLIP-2\end{tabular}  \\ 
		NuScenes-QA\cite{qian2024nuscenes}  & AAAI 2024  & AV(P1) & $\surd$  & $\surd$ & $\surd$       &  & $\surd$ & & MCAN \\
		LLM-Driver\cite{chen2024driving} & ICRA 2024& AV(P3) & & & $\surd$ & $\surd$ & $\surd$ &  & LlaMA-7b \\ 
		PromptTrack\cite{wu2025language} & AAAI 2025 & AV(P1) & $\surd$ &  & $\surd$ &  & $\surd$ &  & PF-Track \\ 
		V2V-LLM\cite{chiu2025v2v} & arXiv 2025 & CAV(V2V-P1P3)  &  & $\surd$ & $\surd$ & $\surd$ & $\surd$ &  & LlaVA-v1.5-7b  \\ \hline
		Talk2DM [this] & --   & CAV(V2C-QP3) & &  & $\surd$ & $\surd$ &  & $\surd$ & \begin{tabular}[c]{@{}c@{}} gpt-oss, deepseek-r1, \\ qwen3, magistral, \\ llama3.1, gemma3\end{tabular} \\ \hline \hline
	\end{tabular}
	\begin{tablenotes}
		\footnotesize
		\item C: Control; 3P: Perception (P1), Prediction (P2), Planning (P3); Q: Query (OTH and NLS querying); V2C: vehicle-to-cloud.
	\end{tablenotes}
\end{threeparttable}
\end{table*}
  
\subsection{Cooperative Perception}
Cooperative perception utilizes wireless communication to exchange sensing data across vehicles and infrastructures, expanding FoV of CAVs. This concept firstly appeared in the grand cooperative driving challenge where participants competed in cooperative driving tasks, such as platoon driving in 2011 \cite{kianfar2012design} and cooperative merging in 2016 \cite{xu2018system}. Contemporary CP strategies consist of three patterns \cite{Huang2025vehicle}: early, intermediate and late cooperation. Early cooperation fuses raw sensor data obtained by agents (vehicles and infrastructures), such as point cloud \cite{chen2023co}. However, sharing raw sensor data needs substantial communication bandwidth. Intermediate cooperation shares middle features that are pre-processed by agents, like BEV features \cite{xu2022cobevt}. Late cooperation that shares perceptual outcomes at object-level is prevailing \cite{tedeschini2023cooperative}, owning to the limited communication capabilities and complexity of feature extraction.
Among these studies, V2V-based architecture is prevailing \cite{xu2018system, xu2022cobevt}, while placing excessive workload on local vehicles potentially impairs the real-time processing capacities of their systems. As wireless communication and cloud computing technologies evolve rapidly, distributing heavy workload to edge and cloud side becomes a more cost-effective and promising solution \cite{tao2024enhancing, gao2024vehicle}. 

\subsection{Dynamic Maps}
The concept of DM stems from the Local Dynamic Map (LDM) \cite{etsi2014intelligent} proposed by European Telecommunication Standards Institute (ETSI). It is designed to be a logical data set that overlays dynamic data gathered from multiple sources onto static road maps \cite{watanabe2020dynamicmap}. Instead of relying solely on sensor data from the surrounding environment, DM collects, manages, and utilizes city-scale sensor data and maps through a geographically decentralized and VRC-cooperative architecture, which comprises embedded devices, edge servers, and cloud servers. Experiments show that the distributed VRC-DM system is able to process a large volume of data in real time with high scalability \cite{hosono2021implementation}. GEV is a representation about the CP and CPd data in such a VRC-DM platform \cite{tao2024enhancing}. In this paper, we focus on the CP data within DM.

\section{System Architecture}
\begin{figure*}[t]
	\centering
	\includegraphics[scale=0.63]{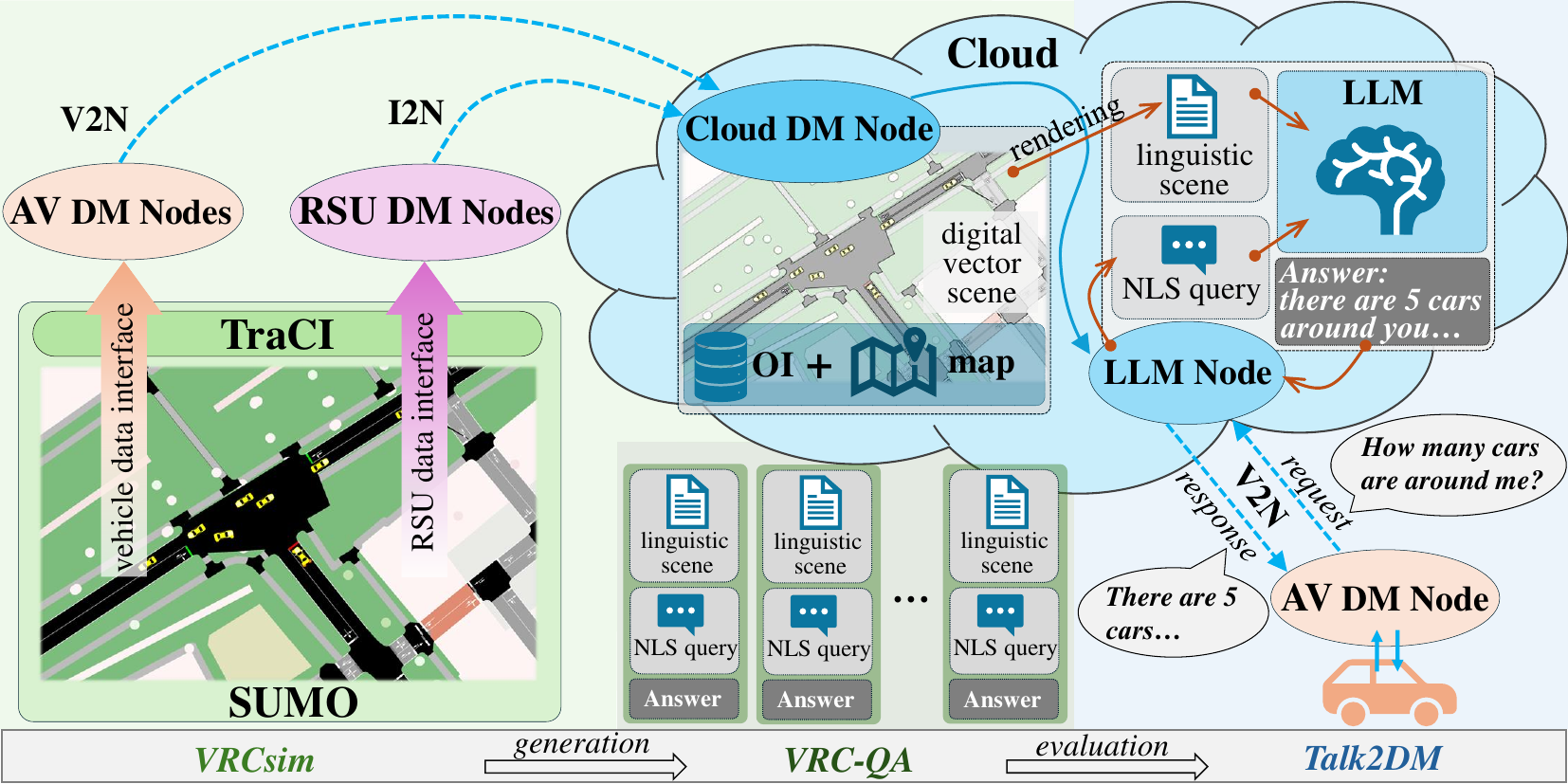}
	\caption{The system architecture. The blue arrows indicate communication connections; the orange arrows represent the data processing flow.}
	\label{fig-archi}
\end{figure*}

In this section, three key components of our system are introduced, including \textbf{\textit{VRCsim}} for simulating VRC cooperation, \textbf{\textit{Talk2BEV}} that enables NLS querying and commonsense reasoning within VRC-DM platforms, and \textbf{\textit{VRC-QA}} dataset for LLMs capacity evaluations. The overall architecture  is shown in \cref{fig-archi}.

\subsection{VRCsim}
VRCsim plays two major roles: 1) simulating the CP scenarios in a VRC cooperation framework and generating structured CP data streams; 2) reconstructing digital traffic scenes based on the aggregated CP data and then rendering these scenes to textual representations for LLMs inputs.

\begin{figure}[htpb]
	\centering
	\includegraphics[scale=0.24]{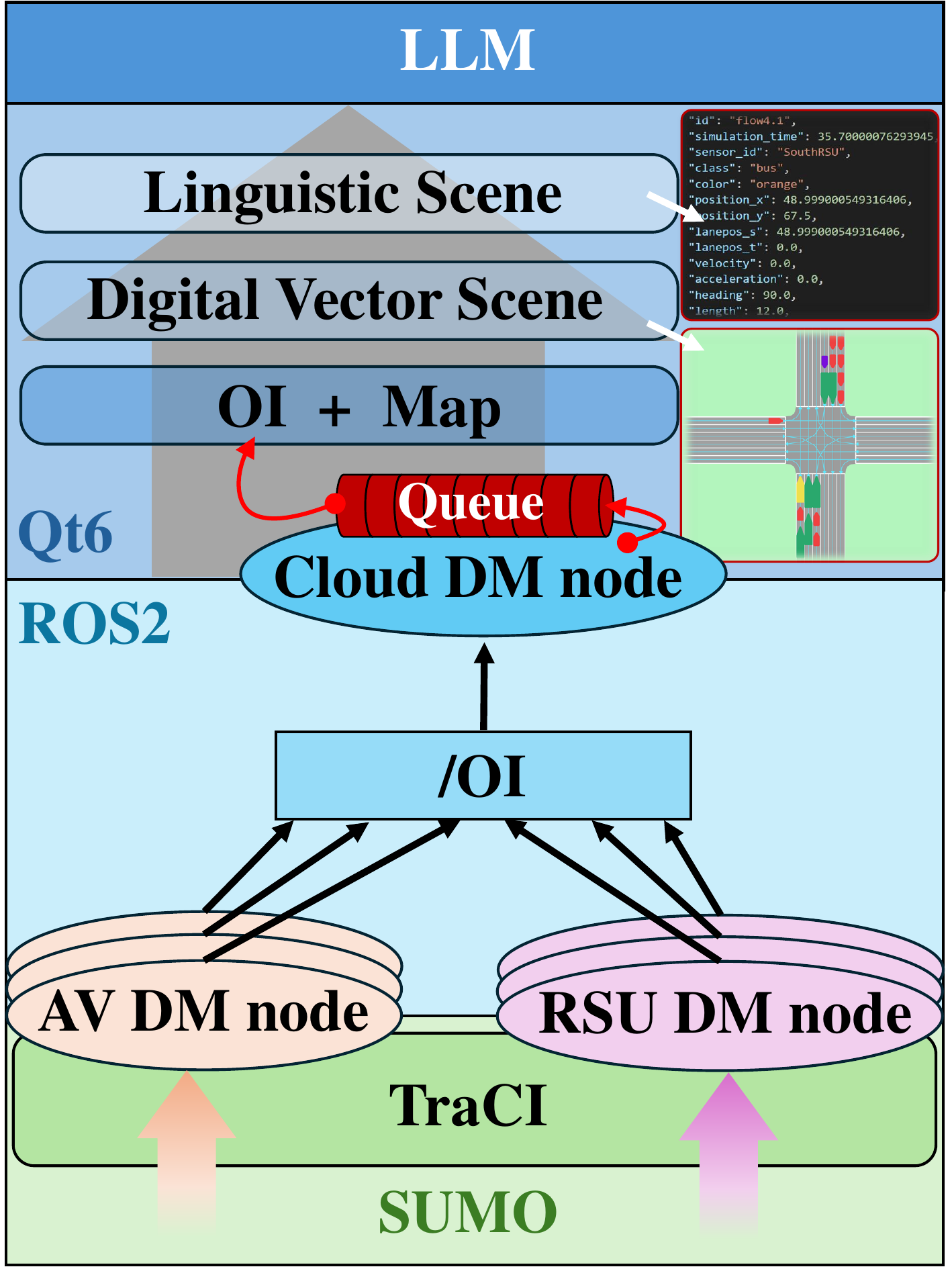}
	\caption{The architecture of VRCsim.}
	\label{fig-vrcsim}
\end{figure}

VRCsim is developed based on three renowned open source platforms: SUMO \cite{lopez2018microscopic}--the microscopic traffic simulation platform, ROS2 \cite{macenski2022robot}--the second generation of the Robot Operating System, and Qt6 \cite{qtweb2025}--the powerful, cross-platform GUI software development framework. The architecture of VRCsim is depicted in \cref{fig-vrcsim}.

In VRCsim, SUMO is used to simulate real-world traffic flows and generate traffic data streams. These data are accessed via Traffic Control Interface (TraCI) \cite{dlrTraCISUMO}, based on which vehicle and RSU data interfaces are established. Using the interfaces, we propose AV and RSU perception models that mimic onboard and roadside sensors, as expressed in equations \cref{eq-vel-model,eq-rsu-model}.
\begin{equation}\label{eq-vel-model}
	P_{av}=\left\{obj \Big| ||(x_{o}, y_{o})-(x_{e}, y_{e})||_2\leq R_s\right\}
\end{equation}
\begin{equation}\label{eq-rsu-model}
	P_{rsu}=\left\{obj \Big| (x_{o}, y_{o}) \in {Ls}, {Ls}=\cup (L_i) \right\}
\end{equation}
where $P_{av}$ and $P_{rsu}$ represent the perception models of AVs and RSUs; $obj$ denotes the perceived object; $(x_{o}, y_{o})$ and $(x_{e}, y_{e})$ are the coordinates of detected objects and ego vehicles, respectively. $R_s$ represents the sensor's detection range of AVs. $L_i$ denotes a lane section on which vehicles are traveling and their movements are captured by roadside sensors.  

Based on \cref{eq-rsu-model,eq-vel-model}, we are able to retrieve associated Object Information (OI, see \cref{tab-oi}) from SUMO using TraCI. The OI from the AV perception model is sent to the AV DM nodes, and the OI from the RSU perception model to the RSU DM nodes. Note that the AV reports the information of both its detected objects and itself. These data are labeled with corresponding detecting sensor ID and then published via the ``/OI'' topic based on ROS2, which builds upon Data Distribution Service (DDS). The ``/OI'' topic is subscribed by a cloud DM node for maintaining the received data. Eventually, we set up a straightforward VRC communication system, simulating the data transmission among a VRC cooperation system like what has been established in \cite{tao2024enhancing}.

\begin{table}[H]
	\caption{Schema of object information retrieved from SUMO.}
	\label{tab-oi}
	\centering
	\begin{tabular}{clcl}
		\hline \hline
		\textbf{Field}  & \textbf{Description}               & \textbf{Field}     & \textbf{Description}          \\ \hline
		id     & identifier                & wi     & width of the object  \\
		t   & time stamp                   & he    & height of the object \\
		x      & x coordinate              & ty      & object type          \\
		y      & y coordinate              & co     & object color         \\
		s      & longitudial lane position & ln      & the current lane     \\
		t      & lateral lane position     & lx     & lane index           \\
		v      & velocity                  & rd      & the current road     \\
		a      & acceleration              & sg    & signal status        \\
		h      & heading                   & ds    & the detecting sensor \\
		le     & length of the object      &       &                      \\ \hline\hline
	\end{tabular}
\end{table}

The streaming perception data--separately obtained by the AV and RSU DM nodes from local areas--are aggregated in the cloud DM node for constructing a more comprehensive traffic scene. A fixed-size queue embedded in the cloud DM node manages these data. To ensure the freshness of received OI, the oldest records are discarded when new data arrive and the queue reaches its capacity. A scene constructor is designed to pull OI from the queue at a predefined frequency (e.g., 5 Hz) to build a comprehensive digital vector scene (DVS) upon on a pre-established static HD map. For the sake of efficiency, the scene constructor is decoupled from DVS, which can then be rendered into both visual and linguistic representations in any desired scale. However, as an initial attempt, this study focuses only on the linguistic scene ($LS$). The $LS$ is transmitted to the downstream LLM node for scene understanding and reasoning using JSON format, as shown in \cref{fig-vrcsim}.
Totally, we generate 12,094 $LS$s for following VRC-QA generation. 

\subsection{VRC-QA}
VRC-QA is a primary contribution of this study. To the best of our knowledge, it is the first VRC-CP dataset that simultaneously involves multiple RSUs and AVs, specially tailored for LLM-AD in VRC cooperative driving scenarios. The procedure of VRC-QA dataset generation is as follows.

\subsubsection{Object Association and Relation Calculation}
given a linguistic scene,
\begin{equation}\label{eq-ls}
	LS=\{o_i, e_j\}, i=1,\cdots,N; j=1,\cdots,M.
\end{equation}
We firstly identify the AV entities $e_j$ and its associated object entities $o_i$ based on their spatial and topological relations. The AV can be easily identified by the id field, which is marked with keyword ``AV''. For example, the id of an AV may be ``AV001''. We then associate the AV entities $e_j$ with surrounding objects $o_i$ based on spatial and topological relations that underpin human's spatial cognition. 

To express the relations among the entities in $LS$, four spatial relations ($R_S$)--front, rear, left, and right--are defined in this paper. Given a AV OI record:
\begin{equation}\label{eq-ego}
	e=[x_{e}, y_{e}, h_{e}, ln_{e}, lx_{e}, rd_{e}, \cdots]
\end{equation}
and a random object OI record:
\begin{equation}\label{eq-obj}
	o=[x_{o}, y_{o}, h_{o}, ln_{o}, lx_{o}, rd_{o}, \cdots]
\end{equation}
the direction angle $\theta$ between them can be calculated by:
\begin{equation}\label{eq-theta}
	\theta=\arccos\frac{(x_{o}-x_{e})\sin h_{e} + (y_{o}-y_{e})\cos h_{e}}{\sqrt{(x_{o}-x_{e})^2 + (y_{o}-y_{e})^2}},
\end{equation}
and then the spatial relation between the AV and the object is then determined through:
\begin{equation}\label{eq-rs}
	R_S=\left\{
	\begin{array}{lr}
		front, &  \theta \in (-45^\circ, 45^\circ]\\
		rear, & \theta \in (135^\circ, 180^\circ] \cup (-180^\circ, -135^\circ]\\
		left, & \theta \in (45^\circ, 135^\circ]\\
		right, & \theta \in (-135^\circ, -45^\circ]\\
	\end{array}
	\right.
\end{equation}

Besides, we also define three lane topology relations ($R_L$): on the left lane, on the right lane, and on the same lane, via:
\begin{equation}\label{eq-lt}
	R_L=\left\{
	\begin{array}{ll}
		leftlane, & rd_{e}=rd_{o} \wedge lx_{e}-lx_{o}=1\\
		rightlane, & rd_{e}=rd_{o} \wedge lx_{o}-lx_{e}=1\\
		samelane, & rd_{e}=rd_{o} \wedge lx_{e}=lx_{o}\\
	\end{array}
	\right.
\end{equation}
and one road topology relation ($R_R$): driving on road $\langle roaddname \rangle$:
\begin{equation}\label{eq-rt}
	R_R=roadname, roadname \rightarrow rd_o.
\end{equation}
$R_S$ and $R_L$ describe the vehicle-to-vehicle relations and $R_R$ presents the vehicle-to-road relation.
Eventually, we can construct a AV-centric entity-relation graph:
\begin{equation}\label{eq-erg}
	G=(LS,R), R=R_S \cup R_L
\end{equation}

It should be noted that a 100-meter spatial limitation is implicitly imposed in the calculation of $G$.

\subsubsection{Attribute-entity-relation Graph Construction}
an entity can be regarded as a collection of attributes. A subset of these attributes is often used for entity identification, while other attributes, possibly outside this subset, constitute the information of interest. The attribute subset de facto corresponds to the observations on which a question is raised regarding the attributes of interest. In other words, humans rely on some attributes to identify the entity of interest, and subsequently query specific attributes associated with it. For example, when we ask: ``\uwave{who} is the \textit{beautiful} \textit{lady} \uline{in front of us} \textit{wearing a red skirt} and \textit{smiling at us}?'', we are in fact using an \textit{attribute set (italicized)} together with a \uline{relation (underlined)} to query an \uwave{unknown attribute of interest (wavy-underlined)}, i.e., the name of the lady. Motivated by this observation, we construct an attribute-entity-relation (AER) graph, as describe below.

\begin{figure}[htpb]
	\centering
	\includegraphics[scale=0.8]{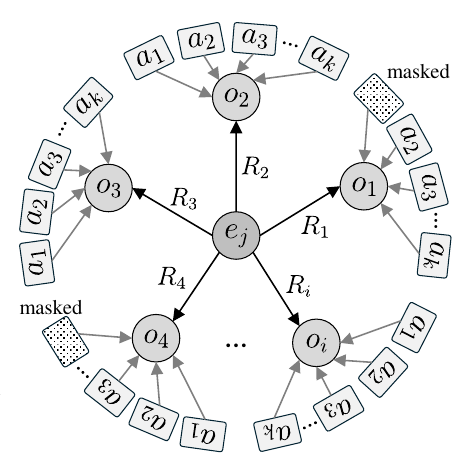}
	\caption{Attribute-entity-relation graph, in which $e_j$ denotes an AV entity (ego vehicle), $o_i$ is an associated object, $a_k$ is an attached attribute, and $R_i$ represents a kind of relations. The attributes and relations are used to generate QA pair.}
	\label{fig-graph}
\end{figure}

Given an $LS$, the attributes listed in \cref{tab-oi} are firstly attached to their corresponding entities. One attribute is then masked (denoted as $a_{masked}$), either randomly or deterministically, and used as the query target. The obtained entity-relation graph in \cref{eq-erg} can be enriched into an AER graph, as shown in \cref{fig-graph}. We construct an ARE graph for each AV in each $LS$.

\subsubsection{QA Template Design and Refinement}

\paragraph{ego-centric QA}
in a given road space, drivers identify entities of interest using the relation relative to the ego vehicle and an attribute set. For example, in the question ``\uwave{how far away} is the \textit{red} \textit{truck} that has \textit{stopped} on the \uline{left lane} from me?'', the attribute sequence $\langle color \rangle$$\langle type \rangle$$\langle velocity \rangle$ along with the relation $\langle leftlane \rangle$ actually points to an entity of interest. 
Suppose there exists a mapping from attributes and relations to objects of interest:
\begin{equation}\label{eq-aro}
	\langle a_{x1}, a_{x2}, \cdots, a_{xi} \rangle \langle R_{xj}\rangle \rightarrow \{o\},
\end{equation}
where $\{o\}$ represents object entities of interest; we can generate QA pairs based on this mapping: 1) selecting a relation from $R$ and an attribute subset from the ARE graph, and fill them into a random question template; 2) assigning the masked attribute (or calculate an attribute, e.g., distance) as the answer. We use the subscripts $xi, xj$ to emphasize that these attributes and relations are variables of unfixed length.

\begin{figure}[htpb]
	\centering
	\includegraphics[scale=0.9]{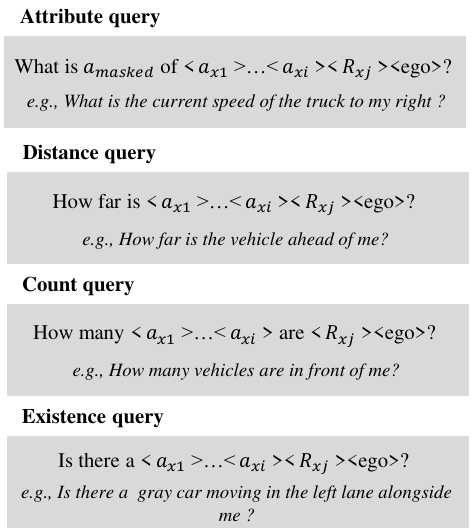}
	\caption{Primitive question templates designed for the ego-centric QA generation. These questions are one-hop.}
	\label{fig-tp-ego}
\end{figure}

Based on \cref{eq-aro}, we design four kinds of primitive question templates, covering attribute, distance, count, and existence queries, as shown in \cref{fig-tp-ego}. They can be instantiated by an attribute subset $\langle a_{x1}, a_{x2}, \cdots, a_{xi} \rangle$ and a relation $\langle R_{xj}\rangle$; and the answers to these questions can be directly obtained from the ARE graph (the masked attributes $a_{masked}$) or computed based on it. \cref{eq-aro} indicates that ego-centric questions are one-hop, since before answering a specific question, we have to identify the relevant entities.

\begin{figure}[htpb]
	\centering
	\includegraphics[scale=0.9]{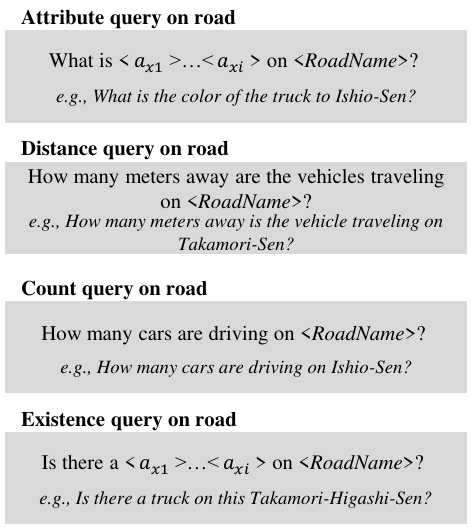}
	\caption{Primitive question templates designed for the ego-agnostic QA generation. These questions are zero-hop.}
	\label{fig-tp-road}
\end{figure}

\paragraph{ego-agnostic QA}
in real world, drivers might be interest in environmental statuses that are indirectly relevant to the ego vehicle, for example the road statuses. We further design four similar primitive question templates, as shown in \cref{fig-tp-road}. Different with ego-centric QA, the ego-agnostic road status QA are zero-hop where objects of interest are associated by $R_R$, which is straightforward. Therefore, the answers of these questions can be directly obtained from the linguistic scene $LS$. 

These manually-designed primitive question templates shown in \cref{fig-tp-ego} and \cref{fig-tp-road} are fed to ChatGPT to enrich their expressing forms. Eventually, we manually check and select 1,121 template variants. Their distribution is expressed in \cref{tab-question-dis}. We denote these question templates as $T_Q$. 

\begin{table}[htpb]
	\caption{Question template Distribution.}
	\centering
	\label{tab-question-dis}
	\begin{minipage}{0.25\textwidth}
		\centering
		\begin{tabular}{p{0.9cm}p{0.9cm}c}
			\hline\hline
			\multicolumn{3}{c}{\textbf{ego-centric}}                                                                         \\ \hline
			\multicolumn{2}{c}{ \textbf{Type}}                                                             & \textbf{Number} \\ \hline
			\multirow{7}{*}{\begin{tabular}[c]{@{}c@{}}attribute\\  query\end{tabular}} 
			& velocity     & 93              \\
			& acceleration & 120              \\
			& heading      & 95               \\
			& color        & 72              \\
			& classification         & 112              \\
			& size         & 96              \\
			& status       & 115             \\
			\multicolumn{2}{c}{distance query}                                                         & 80             \\
			\multicolumn{2}{c}{count query}                                                            & 107              \\
			\multicolumn{2}{c}{existence query}                                                        & 88             \\ \hline
			\multicolumn{2}{c}{total}                                                                  & 978             \\ \hline\hline
		\end{tabular}
	\end{minipage}
	\begin{minipage}{0.25\textwidth}
		\centering
		\begin{tabular}{p{0.9cm}p{0.9cm}c}
			\hline\hline
			\multicolumn{3}{c}{\textbf{ego-agnostic}}                                                                         \\ \hline
			\multicolumn{2}{c}{\textbf{Type}}                                                             & \textbf{Number} \\ \hline
			\multirow{7}{*}{\begin{tabular}[c]{@{}c@{}}attribute\\  query\end{tabular}} 
			& velocity     & 19              \\
			& acceleration & 16              \\
			& heading      & 17               \\
			& color        & 9              \\
			& classification         & 14              \\
			& size         & 16              \\
			& status       & 14             \\
			\multicolumn{2}{c}{distance query}                                                         & 10             \\
			\multicolumn{2}{c}{count query}                                                            & 18              \\
			\multicolumn{2}{c}{existence query}                                                        & 10             \\ \hline
			\multicolumn{2}{c}{total}                                                                  & 143             \\ \hline\hline
		\end{tabular}
	\end{minipage}
\end{table}

\subsubsection{QA Instantiation}
\begin{figure}
	\captionsetup[subfloat]{labelfont=footnotesize, textfont=footnotesize}
	\begin{minipage}[b]{\linewidth}
		\hspace{-0.9cm}
		\subfloat[Question distribution (the left: word count; the right: word cloud).]{
			\includegraphics[scale=1.1]{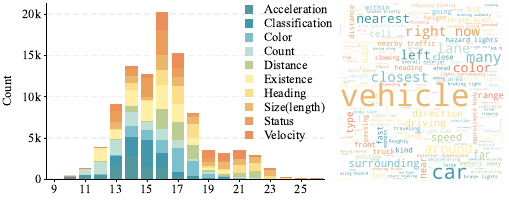}
			\label{fig-question}}
	\end{minipage}
	
	\begin{minipage}[b]{\linewidth}
		\hspace{-0.9cm}
		\subfloat[Answer distribution.]{
			\includegraphics[scale=1.1]{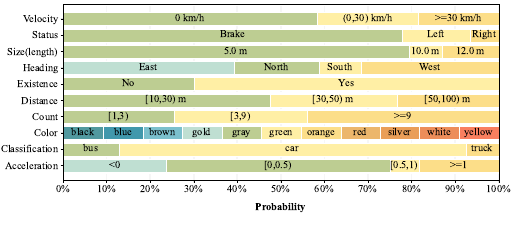}}
		\label{fig-answer}
	\end{minipage}
	\caption{The statistics of VRC-QA dataset.}
	\label{fig-vrc-qa}
\end{figure}

using $T_Q$ and the ARE graph, we can instantiate diverse QA pairs for each $LS$. To create VRC-QA, we randomly select one question template from $T_Q$ to instantiate a QA pair using ARE graph for each $LS$, and repeat this process until 100K QA pairs are obtained. The statistics of VRC-QA is shown in \cref{fig-vrc-qa}.

\subsection{Talk2DM}
Talk2DM is devised to enable NLS querying and commonsense reasoning, serving as a human interface of VRC-DM platforms. In order to maintain the generalization ability of our system and preserve the commonsense reasoning capability of underlying LLMs, Talk2DM is designed as an independent, plug-and-play module built upon open-sourced LLMs, purely based on prompt engineering. 

\begin{figure*}[htpb]
	\centering
	\includegraphics[scale=0.95]{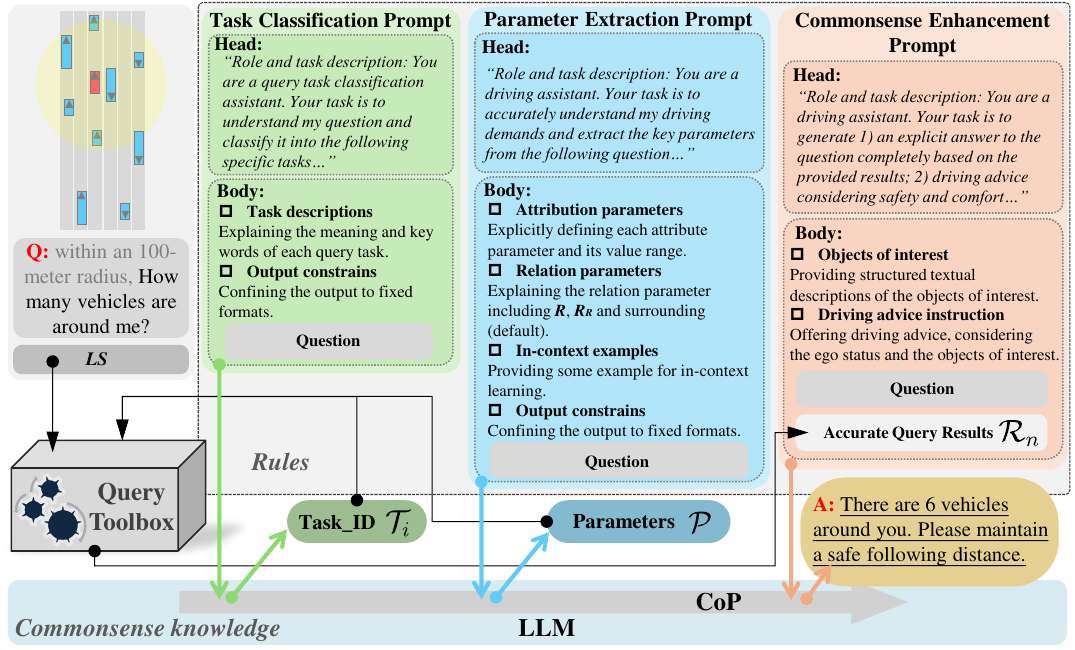}
	\caption{The proposed CoP bridges commonsense knowledge implicit in LLMs and human-defined rules to perform accurate query tasks while offering driving advice. Please note that each question includes a prefixed subtext, ``within an 100-meter radius''. This subtext leads to an abnormal performance degradation in one of our experiments, which we analyze in detail in \cref{sec-cop-qa-s-acc}.}
	\label{fig-cop}
\end{figure*}

To achieve accurate attribute and space querying, we propose a Chain-of-Prompt (CoP, as shown in \cref{fig-cop}) mechanism to guide LLMs to perform accurate NLS querying from structured GEV representations in DM. Essentially, CoP fuses commonsense knowledge implicit in LLMs with human-defined rules, bridging the gap between ambiguous natural language inputs and precise query requirements. This fusion enhances the flexibility of information retrieval in structured database systems. Meanwhile, CoP grounds numerical query results in commonsense knowledge, thereby endowing numeric outputs with interpretable semantic meaning. This capability is crucial for driving decision-making. 

As shown in \cref{fig-cop}, the proposed CoP comprises three prompting phases, progressively ``suturing'' human-defined rules and the commonsense knowledge of LLMs. 
The prompt, in this paper, consists of head and body; the head assigns the role of LLMs and specifies the task to be performed; the body provides detailed information (or rules) to guide the LLMs to complete the tasks, while confining the output to fixed formats to facilitate result processing.
The three CoP phases are detailed below.

\subsubsection{Task Classification Prompting}
to perform precise query task, it is necessary to figure out the potential query type based on the question $Q$. In conventional database system, this process is conducted by human engineers: understanding the query tasks and modeling them through SQL language. In this paper, this process is assigned to LLMs through task classification prompting, as expressed in \cref{eq-tcp}.
\begin{equation}\label{eq-tcp}
	\mathcal{T}_i=\text{LLM}(Q, P_{tc}), \mathcal{T}_i \in T 
\end{equation}
where $P_{tc}$ denotes the task classification prompt; $\mathcal{T}_i$ is the task ID, which is reasoned out based on the question $Q$ from a pre-defined task set $T$. In this paper, $T$ corresponds with the query types exhibited in \cref{tab-question-dis}, containing 10 kinds of query tasks.

As shown in \cref{fig-cop}, $P_{tc}$ instructs LLMs to understand the question and outputs a task ID considering the following task descriptions:
\begin{itemize}
	\item[(1)] speed: querying vehicle speeds, the question may contains keywords like: speed, velocity, how fast, km/h, etc.
	\item[(2)] acceleration: querying vehicle accelerations, the question may contains keywords like: acceleration, deceleration, slowing down, speeding up, etc.
	\item[(3)] heading: querying heading angles of vehicles, the question may contains keywords like: direction, heading, driving toward, etc.  
	\item[(4)] color: querying vehicle colors, the question may contains the keyword: color.
	\item[(5)] classification: querying vehicle types, the question may contains keywords like: type, what type, kind, what kind, what vehicle, etc.
	\item[(6)] size: querying vehicle sizes, the question may contains keywords like: size, length, width, height, how large, how big, body dimension, etc. 
	\item[(7)] status: querying vehicle status, the question may contains keywords like: turn signal, lights, blinker, indicator, signaling, etc.
	\item[(8)] distance: querying the distance between target vehicles and ego vehicle, the question may contains keywords like: distance, how far, how many meters, etc.
	\item[(9)] count: querying the number of target vehicles, the question may contains keywords like: how many, crowded, blocking, dense, etc.
	\item[(10)] existence: querying whether there are vehicles in a certain direction or on a specific lane, the question may contains keywords like: is there, does...exist, is there any, etc. Choose this only when the question is purely about presence/absence and not about lights, signals, distance, color, size, or type.\label{item-existence}
\end{itemize}

\subsubsection{Parameter Extraction Prompting}
in this paper, the accurate query functionality is realized mainly through a set of spatial calculation operators maintained by the query toolbox. The toolbox requires three inputs: a task id $\mathcal{T}_i$, query parameters $\mathcal{P}$ for function calling, and a structured scene representation $LS$. In this phase, LLMs are employed to extract the query parameters from the natural language question $Q$ through parameter extraction prompting, as expressed in \cref{eq-tpe}.
\begin{equation}\label{eq-tpe}
	\mathcal{P}=\text{LLM}(Q, P_{pe})
\end{equation}
where $P_{pe}$ denotes the parameter extraction prompt; the parameter set $\mathcal{P}=\{a_{xi},R_{xi}\}$ serves as relation and attribute constrains that are used to determine function calls in the query toolbox; the logic of the determination is similar to that of \cref{eq-aro}. The attribute parameters mainly include vehicle type and color ; The relation parameters encompass $R$ and $R_R$. If the question $Q$ does not explicitly indicate a relation type in $R \cup R_R$, a default value (surrounding) is assigned. In real world, human typically use these parameters to indicate the objects of interest, for example, 
``the yellow$\langle color \rangle$ truck$\langle type \rangle$ on my left lane $\langle leftlane \rangle$'' or ``the yellow$\langle color \rangle$ truck$\langle type \rangle$ $\langle \text{default:}surrounding \rangle$''. These key parameters are used to identify the objects of interest from a $LS$ by the operators in the query toolbox, and they must firstly be extracted from the question $Q$ prior to function execution. 

Leveraging the powerful reasoning capability, we instruct LLMs to perform parameter extraction using $P_{pe}$, as shown in \cref{fig-cop}. The attribute and relation parameters are explicitly defined in $P_{pe}$ in natural language. In-context learning (ICL) ability is commonly observed in LLMs, pioneering studies \cite{brown2020language, tsimpoukelli2021multimodal} show that feeding in-context examples leads to better few-shot performance. In $P_{pe}$, we include several examples to guide LLMs to perform parameter extraction from questions.

Using the query task id $\mathcal{T}_i$ and query parameters $\mathcal{P}$, accurate calculations can be performed by the toolbox within a specific structured $LS$:
\begin{equation}\label{eq-tb}
	\mathcal{R}_n=\text{ToolBox}(\mathcal{T}_i,\mathcal{P}, LS)
\end{equation}
where $\mathcal{R}_n$ represents the accurate numeric results retrieved from the traffic scenes. Essentially, function ToolBox() is a branching logic that controls the functions calling and execution flow and produces the numeric results $\mathcal{R}_n$.

\subsubsection{Commonsense Enhancement Prompting}
the queried numerical results are insufficient for driving decisions, since they are often inconclusive and lack clear semantic implications. 
Therefore, enriching the query result with semantic driving advice--namely, commonsense enhancement--is crucial. This phase is also perform by LLMs through commonsense enhancement prompting, as expressed in \cref{eq-ce}.
\begin{equation}\label{eq-ce}
	\mathcal{R}_s, \mathcal{V}=\text{LLM}(\mathcal{R}_n, Q, P_{ce})
\end{equation} 
where $P_{ce}$ represents the commonsense enhancement prompt; $\mathcal{R}_s$ is the semantic query results;  $\mathcal{V}$ is the driving advice. The final answer $A$ of question $Q$ equals to $\mathcal{R}_s \oplus \mathcal{V}$.

It is worth noting that in $P_{ce}$, we include the complete OI data (see \cref{tab-oi}) of retrieved objects, which can be regarded as a description of the relevant surrounding environments. $P_{ce}$ instructs LLMs to provide driving advice by jointly considering the information of the ego vehicle and its surrounding environments.

Finally, Talk2DM is established based on the aforementioned CoP, which can be formally expressed by:
\begin{equation}\label{eq-cop}
		A=\text{LLM}(\text{ToolBox}(
		\text{LLM}(Q, P_{tc}),
		\text{LLM}(Q, P_{pe}), LS),Q, P_{ce})
\end{equation}

\section{Results}
\subsection{Experimental Configurations}
\subsubsection{Data}
\begin{figure}[htpb]
	\centering
	\includegraphics[scale=1]{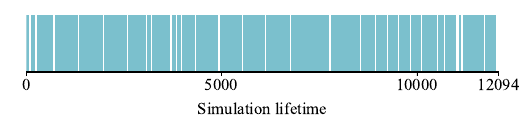}
	\caption{Simulation lifetime and sampled scenes; a blue vertical line is plotted at each sampled timestamp along the simulation lifetime axis (frame id).}
	\label{fig-frames}
\end{figure}

we randomly select 10K QA pairs from VRC-QA to evaluate the NSL querying performance of Talk2DM. These data cover all the query types (see \cref{tab-q-dis-exp}) and span 1,420 VRC cooperation scenes, which are evenly distributed over the entire simulation lifetime (see \cref{fig-frames}). 
\begin{table}[htpb]
\centering

	\caption{distribution of QA pairs used in this experiment.}
	\label{tab-q-dis-exp}
	\begin{tabular}{clc}
		\hline \hline
		\multicolumn{2}{c}{\textbf{Question type}}                                                             & \textbf{Number} \\ \hline
		\multirow{7}{*}{\begin{tabular}[c]{@{}c@{}}attribute\\  query\end{tabular}} 
		& velocity     & 962              \\
		& acceleration & 999              \\
		& heading      & 990               \\
		& color        & 956              \\
		& classification     & 941              \\
		& size         & 959              \\
		& status       & 943             \\
		\multicolumn{2}{c}{distance query}                                                         & 926             \\
		\multicolumn{2}{c}{count query}                                                            & 961              \\
		\multicolumn{2}{c}{existence query}                                                        & 1,363             \\ \hline
		\multicolumn{2}{c}{total}                                                                  & 10,000             \\ \hline \hline
		\end{tabular}
\end{table}

\subsubsection{Models}
Talk2DM is designed as a plug-and-play module independent of underlying LLMs, enabling seamless integration with different models to exploit their commonsense reasoning capabilities. We evaluate six series of reasoning LLMs (17 models in total), whose configurations are reported in \cref{tab-llm-config}.

\newcolumntype{C}{>{\centering\arraybackslash}X}
\begin{table}[htpb]
	\footnotesize
	\centering
\begin{threeparttable}
	\caption{Overview of LLM configurations.}
	\label{tab-llm-config}
	\begin{tabularx}{\columnwidth}{CCCCCC}
		\hline \hline
		\multicolumn{2}{c}{\textbf{Model}} &
		\begin{tabular}[c]{@{}c@{}}\textbf{Archit}-\\ \textbf{ecture}\end{tabular} &
		\begin{tabular}[c]{@{}c@{}}\textbf{Context}\\ \textbf{length}\end{tabular} &
		\begin{tabular}[c]{@{}c@{}}\textbf{Embedding}\\ \textbf{length}\end{tabular} &
		\begin{tabular}[c]{@{}c@{}}\textbf{Quanti}-\\ \textbf{zation}\end{tabular} \\ \cline{1-2}
		
		\textbf{Family} & \textbf{Size} & & & & \\ \hline
		
		GPT-oss & 20B & gptoss & 131072 & 2880 & MXFP4 \\ \hline
		
		\multirow{4}{*}{Deepseek-r1}
		& 1.5B & qwen2 & 131072 & 1536 & Q4\_K\_M \\
		& 8B   & qwen3 & 131072 & 4096 & Q4\_K\_M \\
		& 14B  & qwen2 & 131072 & 5120 & Q4\_K\_M \\
		& 32B  & qwen2 & 131072 & 5120 & Q4\_K\_M \\ \hline
		
		\multirow{6}{*}{Qwen3}
		& 0.6B & qwen3    & 40960  & 1024 & Q4\_K\_M \\
		& 1.7B & qwen3    & 40960  & 2048 & Q4\_K\_M \\
		& 4B   & qwen3    & 262144 & 2560 & Q4\_K\_M \\
		& 8B*   & qwen3    & 40960  & 4096 & Q4\_K\_M \\
		& 14B  & qwen3    & 40960  & 5120 & Q4\_K\_M \\
		& 30B  & qwen3moe & 40960  & 5120 & Q4\_K\_M \\ \hline
		
		Magistral & 24B & llama & 40000 & 5120 & Q4\_K\_M \\ \hline
		
		\multirow{4}{*}{Gemma}
		& 1B  & gemma3 & 32768  & 1152 & Q4\_K\_M \\
		& 4B  & gemma3 & 131072 & 2560 & Q4\_K\_M \\
		& 12B & gemma3 & 131072 & 3840 & Q4\_K\_M \\
		& 27B & gemma3 & 131072 & 5376 & Q4\_K\_M \\ \hline
		
		Llama3 & 8B & llama & 8192 & 4096 & Q4\_0 \\ \hline \hline
	\end{tabularx}
	\begin{tablenotes}
		\footnotesize
		\item *Reference model used in Talk2DM development.
	\end{tablenotes}
\end{threeparttable}
\end{table}

\subsubsection{Metrics}
given a predicted task classification id $\mathcal{T}_i$ and a ground-truth task id $\hat{\mathcal{T}}_i$, the classification accuracy $\mathcal{A}_C$ is defined as:
\begin{equation}\label{eq-ac}
	\mathcal{A}_C = \frac{\sum_{i=1}^{N}C_i}{N_p} \times 100\%
\end{equation}
where $C_i=1$ if $\mathcal{T}_i=\hat{\mathcal{T}}_i$ and $C_i=0$ otherwise; $N_p$ denotes the total number of predictions. $\mathcal{A}_C$ reflects the query task comprehension capacities of LLMs. 

Further, suppose the numeric query results are denoted by $\mathcal{R}_n=[r_1,r_2,\cdots,r_k]$ and the corresponding ground-truth results by $\hat{\mathcal{R}}_n=[\hat{r}_1,\hat{r}_2,\cdots,\hat{r}_k]$, we consider $\hat{\mathcal{R}}_n$ to be correct if:
\begin{equation}\label{eq-rc}
	\mathcal{R}_n=\hat{\mathcal{R}}_n \iff \forall i=1,\cdots,k, r_i=\hat{r}_i.
\end{equation}
Given a query task $\mathcal{T}_i$, the query accuracy performance of model $\mathcal{M}_k$ is defined as:
\begin{equation}\label{eq-aq}
	\mathcal{A}_Q^{\mathcal{M}_k}(\mathcal{T}_i)=\frac{N_c}{N_q}\times 100\%,
\end{equation}
where $N_c$ is the total number of correct queries and $N_q$ is the total number of queries. $\mathcal{A}_Q^{\mathcal{M}_k}$ reflects the query accuracy of Talk2DM using model $\mathcal{M}_k$, implicitly capturing the commonsense reasoning capacities of LLMs .

To access the task-specific performance bias, we define the following task bias indicator:
\begin{equation}\label{eq-bias}
	\mathcal{B}_{\mathcal{T}_i}^{\mathcal{M}_k} = \left\{
	\begin{array}{cl}
		+1, & \text{if } \mathcal{A}_Q^{\mathcal{M}_k}(\mathcal{T}_i) \geq \mathcal{A}_Q^{\mathcal{M}_k}(\mathcal{T}_j), i\neq j;\\
		-1, & \text{if } \mathcal{A}_Q^{\mathcal{M}_k}(\mathcal{T}_i) < \mathcal{A}_Q^{\mathcal{M}_k}(\mathcal{T}_j), i\neq j;\\
		0, &\text{else.}
	\end{array}
	\right.
\end{equation}
The task-specific performance bias of Talk2DM on a specific query task is:
\begin{equation}
	\mathcal{B}_{\mathcal{T}_i} = \sum \mathcal{B}_{\mathcal{T}_i}^{\mathcal{M}_k},
\end{equation}
which reflect the task-specific performance bias of Talk2DM.


\subsubsection{Experiments}
in this paper, we design three experiments to evaluate Talk2DM in collaboration with different LLMs:
\begin{itemize}
	\item OSP-QA: one-shot prompting-based QA, which performs question-answering using a single prompting step; such strategies have been adopted in classical VQA applications\cite{yang2022empirical, guo2023from,shao2023prompting};
	\item CoP-QA-F: CoP-based QA, which applies the proposed CoP strategy to perform QA tasks with different LLM families (F); this experiment aims to identify the strongest model in terms of NLS querying and commonsense reasoning;
	\item CoP-QA-S: CoP-based QA, which applies the proposed CoP strategy to conduct QA tasks with different model sizes (S); this experiment aims to identify the most appropriate model by jointly considering query accuracy and efficiency.
\end{itemize}
All experiments are conducted locally on an RTX 5090 D GPU with 32GB of memory. We use the same CoP $\{P_{tc},P_{pe},P_{ce}\}$ when integrating different underlying LLMs; these prompts are developed and iteratively refined using Qwen3:8B model as the reference model; they are then directly applied to other models without further modification to examine the generalization ability of Talk2DM.

\subsection{OSP-QA}
In this experiment, we compare the NLS querying performance of the proposed CoP with the classical one-shot prompting (OSP) method. We design four OSP approaches:
\begin{itemize}
	\item OSP1: the prompt only includes role setting, \textit{LS}, road information, and output constrains;
	\item OSP2: OSP1 + explanations of OI (\cref{tab-oi});
	\item OSP3: OSP1 + inference rules (defined by us);
	\item OSP4: OSP1 + few-shot examples.
\end{itemize}
They are evaluated on VRC-QA and compared with Talk2DM employing the CoP. Both the SOP methods and Talk2DM use the Qwen3:8B model as underlying the LLM. The results are presented in \cref{tab-sop-qa}.

\begin{table}[htpb]
\centering
	\begin{threeparttable}
	\caption{NLS query accuracies of OSP methods and Talk2DM.}
	\label{tab-sop-qa}
	\begin{tabular}{lccccc}
		\hline \hline
		Query task / method & OSP1  & OSP2  & OSP3  & OSP4  & Talk2DM \\ \hline
		(1) velocity       & 37.86 & 36.89 & 40.78 & 35.56 & 96.81   \\
		(2) acceleration   & 15.79 & 18.95 & 16.84 & 17.86 & 86.27   \\
		(3) heading        & 32.35 & 34.31 & 32.35 & 37.21 & 87.50   \\
		(4) color          & 33.70 & 33.70 & 39.13 & 41.03 & 99.06   \\
		(5) classification & 29.79 & 41.49 & 42.55 & 43.21 & 96.60   \\
		(6) size           & 32.95 & 23.86 & 27.27 & 28.21 & 97.82   \\
		(7) status         & 33.67 & 34.69 & 30.61 & 40.70 & 89.34   \\
		(8) distance       & 1.11  & 2.22  & 1.11  & 1.28  & 97.73   \\
		(9) count          & 45.00 & 47.00 & 46.00 & 42.22 & 96.70   \\
		(10) existence      & 74.64 & 83.33 & 71.74 & 81.51 & 95.55   \\ \hline
		\textbf{Average}        & 33.69 & 35.65 & 34.84 & 36.88 & 94.34   \\ \hline \hline
	\end{tabular}
	\begin{tablenotes}
		\footnotesize
		\item $\mathcal{A}_Q$, unit: percent \%.
	\end{tablenotes}
\end{threeparttable}
\end{table}

The results in \cref{tab-sop-qa} clearly show that Talk2DM significantly outperforms the classical OSP methods. The best-performing OSP4 method achieves only 36.88\% query accuracy, which lags far behind Talk2DM, attaining 94.34\%. Generally, OSP methods severely botch space-related NLS queries, as evidenced by their poor performance, particularly on distance queries.

A noticeable exception is the existence query, for which OSP methods achieve  results comparable to those of Talk2DM. In fact, LLMs do not truly reason over this kind of questions; instead, they largely rely on guessing based on the input $LS$. Since existence queries are binary (yes or no), the limited output space inflates the apparent accuracy. Consequently, high accuracy can be obtained even through no actual spatial computation is performed by the OSP methods.   

\subsection{CoP-QA-F}
This experiment demonstrates the generalization ability of Talk2DM and compares NLS querying performance across different LLM families. We choose the largest models of different LLM families executable on our hardware settings. The overall results are presented in \cref{tab-cop-f}.

The experimental results shows that Talk2DM can be effortlessly adopted to various LLMs from Qwen3 family, such as GPT-oss and Gemma3 families, while maintaining competitive performance. Qwen3:30B model achieves the highest query accuracy $\mathcal{A}_Q=95.15\%$, followed by GPT-oss:20B model with $\mathcal{A}_Q=94.45\%$, and then Gemma3:27B model with $\mathcal{A}_Q=93.09\%$. It is worth noting that Gemma3:27B achieves the best result in query task classification with $\mathcal{A}_C=97.60\%$, better than Qwen3:30B ($\mathcal{A}_C=96.55\%$) model which severs as the reference model in Talk2DM development. These results demonstrate the robust generalization ability of Talk2DM.

The Qwen3:30B, GPT-oss:20B and Gemma3:27B models demonstrate competitive performance in query demand comprehension and commonsense reasoning, preserving high levels of both $\mathcal{A}_C$ and $\mathcal{A}_Q$ (above 93\%). The Magistral:24B model performs well in understanding query demands, with $\mathcal{A}_C=90.81\%$, but exhibits poor performance in query parameter extraction, likely due to its relatively weak commonsense reasoning capacity, resulting in a low query accuracy of $\mathcal{A}_Q=71.78\%$. The Deepseek-r1:32B model, which shares the same architecture and size as Qwen3:30B, shows a significant accuracy gap compared to Qwen3:30B. The Llama3.1:8B model performs the worst among these LLMs, possibly because of its small model size. Notably, the Magistral:24B model, despite using the same llama architecture with a larger 24B size, still does not perform well in NLS querying.

\cref{tab-cop-f} shows that Talk2DM struggles with acceleration ($\mathcal{B}_{\mathcal{T}_2}=-2$) and heading ($\mathcal{B}_{\mathcal{T}_3}=-2$) related queries, while performing well on color ($\mathcal{B}_{\mathcal{T}_4}=+4$) and classification ($\mathcal{B}_{\mathcal{T}_5}=+2$) queries. This discrepancy primarily stems from ambiguities in natural language questions. For example, questions such as ``\textit{is the bus close to my car keeping a constant speed?}'' and ``\textit{is the truck near me suddenly slowing down?}'', implicitly require inferring vehicle acceleration rather than retrieving speed. This literal ambiguity encourages surface-level reasoning, leading to failures on acceleration queries. The negative performance bias of acceleration query is commonly observed in Qwen3 ($\mathcal{B}_{\mathcal{T}_2}=-4$, see \cref{tab-cop-s-qwen}), Deekseek-r1 ($\mathcal{B}_{\mathcal{T}_2}=-2$, see \cref{tab-cop-s-deepseek}) and Gemma3 ($\mathcal{B}_{\mathcal{T}_2}=-2$, see \cref{tab-cop-s-gemma}).
Besides, Talk2DM using the Llama3.1:8B model underperforms on heading queries, as nearly 80\% of them are misclassified into other query types. For instance, the question ``\textit{where is the vehicle ahead currently driving towards?}'' is incorrectly interpreted as an acceleration-related query by Llama3.1:8B model. This phenomenon appears to be model-specific, as it is not observed in other models reported in \cref{tab-cop-s-qwen}-\cref{tab-cop-s-gemma}.    
In contrast, color queries are typically explicit and unambiguous (e.g., ``\textit{what is the color of the bus around me?''} and ``\textit{what color does the vehicle around my car appear to be?}''), and therefore achieve strong performance. These observations indicate that, when faced with ambiguous or implicit natural language questions, LLMs still lack sufficient depth of reasoning.

\begin{table*}[htpb]
	\centering
\begin{threeparttable}
	\caption{Experimental results of CoP-QA-F.}
	\label{tab-cop-f}
	\begin{tabular}{lccccccccccccr}
		\hline \hline
		\textbf{Talk2DM using:}  & \multicolumn{2}{c}{Qwen3:30B} & \multicolumn{2}{c}{GPT-oss:20B} & \multicolumn{2}{c}{Magistral:24B} & \multicolumn{2}{c}{Deepseek-r1:32B} & \multicolumn{2}{c}{Gemma3:27B} & \multicolumn{2}{c}{Llama3.1:8B} & \multirow{2}{*}{$\mathcal{B}_{\mathcal{T}_i}$} \\ \cline{2-13}
		 \textbf{Query task} & $\mathcal{A}_C$                                 & $\mathcal{A}_Q$                                 & $\mathcal{A}_C$                                  & $\mathcal{A}_Q$                                  & $\mathcal{A}_C$                                   & $\mathcal{A}_Q$                                    & $\mathcal{A}_C$                                    & $\mathcal{A}_Q$                                     & $\mathcal{A}_C$                                  & $\mathcal{A}_Q$                                 & $\mathcal{A}_C$                                  & $\mathcal{A}_Q$      &                                                           \\ \hline
		(1) velocity                                                              & 99.01                              & 97.03                              & 99.48                               & 96.59                               & 83.77                                & 66.33                                & 78.92                                & \uline{77.45}                                 & 96.75                               & 94.12                              & 36.31                               & 34.48    & -1    \\
		(2) acceleration                                                          & 76.60                             & \uline{75.53}                              & 90.05                               & 88.36                               & 66.00                                & \uline{50.70}                                & 82.83                              & 81.31                                 & 93.56                               & 91.95                              & 57.55                               & 54.53        & -2                \\
		(3) heading                                                               & 100                                & 92.08                              & 98.30                               & \uline{87.90}                               & 84.16                                & 62.77                                & 92.20                                 & 85.37                                 & 98.22                               & 92.67                              & 20.59                              & \uline{15.05} & -2    \\
		(4) color                                                                 & 100                                & \uwave{100}                                & 100                                 & \uwave{99.79}                               & 100                                  & \uwave{86.29}                                & 90.16                                 & 89.62                                 & 100                                 & \uwave{100}                                & 99.79                               & 87.55                  & +4             \\
		(5) classification                                                        & 100                                & \uwave{100}                                & 100                                 & 97.47                               & 98.98                                & 75.36                                & 87.88                                 & 87.88                                 & 100                                 & 84.11                              & 99.80                               & \uwave{91.65}          & +2                     \\
		(6) size                                                                  & 98.85                              & 98.85                              & 100                                 & 99.69                               & 98.95                                & 72.38                                & 92.89                                 & \uwave{91.88}                                 & 100                                 & 93.72                              & 100                                 & 87.24                    & +1           \\
		(7) status                                                                & 96.88                              & 94.79                              & 90.93                               & 89.87                               & 93.19                                & 75.96                                & 69.70                                 & 68.18                                 & 90.00                              & \uline{83.83}                              & 96.38                               & 85.74                & -1               \\
		(8) distance                                                              & 100                                & \uwave{100}                                & 100                                 & 99.03                               & 96.23                                & 74.28                                & 91.71                                 & 90.61                                 & 100                                 & 98.23                              & 90.91                               & 82.26              & +1                 \\
		(9) count                                                                 & 100                               & 99.05                              & 99.07                               & 96.59                               & 99.38                                & 80.21                                & 88.89                                 & 87.88                                 & 99.18                               & 96.49                              & 98.56                               & 89.90                     & 0          \\
		(10) existence                                                             & 94.16                              & 94.16                              & 90.28                               & 89.19                               & 87.48                                & 73.51                                & 93.29                                 & 91.52                                 & 98.25                               & 95.78                              & 91.70                               & 86.17               & 0                \\ \hline
		\textbf{Average}                                                               & 96.55                              & \uwave{95.15}                              & 96.81                               & 94.45                               & 90.81                                & 71.78                                & 86.85                                 & 85.17                                 & 97.60                               & 93.09                              & 79.16                              & \uline{71.46}        & --                       \\ \hline \hline
	\end{tabular}
	\begin{tablenotes}
		\footnotesize
		\item \uline{minimum $\mathcal{A}_Q$}, \uwave{maximum $\mathcal{A}_Q$}.
	\end{tablenotes}
\end{threeparttable}
\end{table*}

\subsection{CoP-QA-S}
\subsubsection{Accuracy}\label{sec-cop-qa-s-acc}
\begin{figure}[htpb]
	\centering
	\includegraphics[scale=0.8]{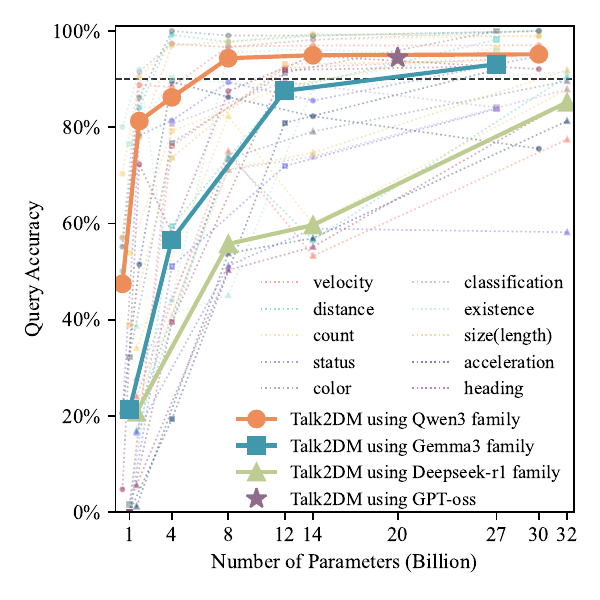}
	\caption{Query accuracy of the Qwen3, Deepseek-r1, Gemma3 model families. The solid lines denote the average query accuracy across ten types of query tasks, while the dotted lines represent the query accuracy for individual query types.}
	\label{fig-cop-qa-s}
\end{figure}

In this experiment, we evaluate Talk2DM's query accuracy across different model sizes, ranging from 0.6B to 32B, using the Qwen3, Deepseek-r1, and Gemma3 model families. Detailed model configurations have been summarized  in \cref{tab-llm-config}. The overall results of the CoP-QA-S experiment are reported in \cref{tab-cop-s-qwen}, \cref{tab-cop-s-deepseek} and \cref{tab-cop-s-gemma} for the Qwen3, Deepseek-r1, and Gemma3 model families, respectively. For clarity, all these results are further visualized in \cref{fig-cop-qa-s}.

\cref{fig-cop-qa-s} clearly demonstrates that the average query accuracy of Talk2DM consistently improves as model size increases, across all model families. The Qwen3 family outperforms the Deepseek-r1 and Gemma3 families, primarily because the Qwen3:8B model was used as the reference model during Talk2DM development. The Gemma3 family and GPT-oss:20B model achieve performance comparable to that of the Qwen3 family as model size increases, even though no modifications are applies to the CoP when switching the underlying LLMs from Qwen3:8B to Gemma3 and GPT-oss models. A similar trend is observed for the Deepseek-r1 family; however, a clear performance gap remains between it and the other three families.

\begin{table*}[htpb]
	\centering
	\caption{Experimental results of CoP-QA-S using Qwen3 family.}
	\label{tab-cop-s-qwen}
	\begin{tabular}{lccccccccccccr}
		\hline\hline
		\textbf{Talk2DM using} & \multicolumn{2}{c}{Qwen3:0.6B} & \multicolumn{2}{c}{Qwen3:1.7B} & \multicolumn{2}{c}{Qwen3:4B} & \multicolumn{2}{c}{Qwen3:8B*} & \multicolumn{2}{c}{Qwen3:14B} & \multicolumn{2}{c}{Qwen3:30B} & \multirow{2}{*}{$\mathcal{B}_{\mathcal{T}_i}$} \\ \cline{2-13} 
		\textbf{Query task}    & $\mathcal{A}_C$             & $\mathcal{A}_Q$            & $\mathcal{A}_C$             & $\mathcal{A}_Q$            & $\mathcal{A}_C$            & $\mathcal{A}_Q$           & $\mathcal{A}_C$            & $\mathcal{A}_Q$           & $\mathcal{A}_C$            & $\mathcal{A}_Q$            & $\mathcal{A}_C$            & $\mathcal{A}_Q$            \\ \hline
		(1) velocity               & 96.55          & 48.88         & 99.63          & 88.76         & 99.25         & 88.39        & 99.59         & 96.81        & 99.70         & 96.96         & 99.01         & 97.03        & 0 \\
		(2) acceleration           & 26.36          & 20.52         & 59.02          & \uline{51.50}         & 91.73         & 89.47        & 88.56         & \uline{86.27}        & 84.08         & \uline{82.28}         & 76.60         &\uline{75.53}    &  -4  \\
		(3) heading                & 8.71           & \uline{4.75}          & 89.21          & 72.30         & 97.84         & 67.99        & 97.80         & 87.50        & 95.69         & 94.83         & 100           & 92.08   & -1      \\
		(4) color                  & 95.36          & 65.19         & 100            & 86.18         & 100           & \uwave{100}          & 99.89         & \uwave{99.06}        & 99.04         & 99.04         & 100           & \uwave{100 }     &  +3   \\
		(5) classification         & 86.56          & 57.03         & 99.26          & 91.52         & 100           & 97.42        & 99.59         & 96.60        & 99.70         & 98.19         & 100           & \uwave{100}      & +1     \\
		(6) size                   & 88.70          & 57.11         & 97.71          & 77.86         & 100           & 96.95        & 99.06         & 97.82        & 99.69         & \uwave{99.39}         & 98.85         & 98.85    & +1     \\
		(7) status                 & 31.06          & 20.64         & 93.80          & 78.29         & 88.76         & 81.40        & 90.82         & 89.34        & 85.53         & 85.53         & 96.88         & 94.79     & 0    \\
		(8) distance               & 67.40          & 50.11         & 95.49          & 84.02         & 100           & 99.18        & 100           & 97.73        & 99.67         & 99.00         & 100           & \uwave{100}     & +1      \\
		(9) count                  & 94.85          & 70.31         & 98.85          & 90.38         & 99.62         & 97.31        & 98.86         & 96.70        & 99.09         & 98.17         & 100           & 99.05    & 0     \\
		(10) existence              & 97.53          & \uwave{80.06}         & 96.53          & \uwave{92.00}         & 44.50         & \uline{44.00}        & 97.37         & 95.55        & 97.87         & 96.17         & 94.16         & 94.16    & +1     \\ \hline
		\textbf{Average}       & 69.31          & 47.46         & 92.95          & 81.28         & 92.20         & 86.21        & 97.15         & 94.34        & 96.01         & 94.96         & 96.55         & 95.15   & --     \\ \hline\hline
	\end{tabular}
\end{table*}

\cref{tab-cop-s-qwen} presents the detailed experimental results of Qwen3 model family across each query task. Averagely, the querying performance improves as the model size increases, achieving the best performance of 95.15\% with 30B model. However, due to the reference model setting of Qwen3:8B, we can only observe marginal performance improvements using 14B and 30B models. 

Among the Qwen3 family experiments, an abnormal and sharp drop is observed on existence query tasks when using the 4B model with $\mathcal{A}_Q(\mathcal{T}_{10})$ decreasing from 92\% to 44\% (with $\mathcal{A}_C=44.5\%$) as the model size increases from 1.7B to 4B. To figure out the reason, we carry out an ablation experiment and present the results in \cref{tab-ablation-sr}. 
\begin{table*}[htpb]
	\centering
	\begin{threeparttable}
		\caption{Ablation analysis of the sharp drop in existence-query performance with Qwen3:4B.}
		\label{tab-ablation-sr}
		\begin{tabular}{cccccccccccc}
			\hline \hline
			Task classification:& velocity$\downarrow$ & acceleration$\downarrow$ & heading$\downarrow$ & color$\downarrow$ & classification$\downarrow$ & size$\downarrow$ & status$\downarrow$ & distance$\downarrow$ & count$\downarrow$ & none$\downarrow$ & existence $\uparrow$ \\ \hline
			\multicolumn{1}{l|}{with \textbf{s \& r}}    & 4.83              & 2.15                  & 1.88             & 0.27           & 0.80                    & 0             & 1.07            & 43.43             & 1.07           & 0             & 44.50              \\
			\multicolumn{1}{l|}{remove \textbf{s}}      & 1.61              & 4.02                  & 2.68             & 0              & 1.07                    & 0.27          & 0.27            & 11.26             & 0              & 0.80          & 78.02              \\
			\multicolumn{1}{l|}{remove \textbf{s \& r}} & 0                 & 0                     & 0                & 0              & 0.80                    & 0             & 0               & 0                 & 0.27           & 0             & 98.93              \\ \hline\hline
		\end{tabular}
		\begin{tablenotes}
			\footnotesize
			\item unit: percent \%; this table reports the proportion of existence questions classified into different query types; for example, the value of 4.83 indicates that 4.83\% of existence questions are misclassified as velocity queries; the existence column reflects the correct task classification rate; the none column indicate classification failure rate.
		\end{tablenotes}
	\end{threeparttable}
\end{table*}
This table shows that the drop is mainly caused by 1) the prefixed subtext of each question, ``\textit{within an 100-meter radius}'' (denoted as \textbf{s}); and 2) the restrictive rule (refer to \cref{item-existence}) for existence query in task classification prompt $P_{tc}$, ``\textit{Choose this only when the question is purely about presence/absence and not about lights, signals, distance, color, size, or type}'' (denoted as \textbf{r}). The subtext \textbf{s} confuses the Qwen3:4B model, causing 43.43\% of existence queries to be misclassified as distance queries. After removing \textbf{s}, this rate is reduced to 11.26\%. Although replacing the subtext with a fixed parameter could potentially improve the overall performance, we retain it to preserve the flexibility of Talk2DM. The restrictive rule \textbf{r} was originally designed to prevent the misclassification of existence queries and functions well with Qwen3:8B and other models; however, it exhibits a negative effect when the Qwen3:4B model is used. Ultimately, removing both \textbf{s} and \textbf{r} improves the classification accuracy of existence queries from 44.8\% to 98.93\%. This ablation experiment reveals that well-tested prompts do not always perform well on specific tasks; however, we can also observe the generalization ability of our CoP when applied to the Qwen3:4B and Qwen3:1.7B models, maintaining relatively high query accuracies of 81.28\% and 86.21\%, respectively.

The experimental results for Gemma3 and Deepseek-r1 families are respectively summarized in \cref{tab-cop-s-gemma} and \cref{tab-cop-s-deepseek}. Similar to Qwen3 family, larger models consistently achieve better performance. The  Gemma3:12B and Gemma3:27B models preserve strong performance when directly used in Talk2DM, achieving query accuracies of 87.56\% and 93.09\%, respectively, which are comparable to those of the Qwen3:4B-30B models. However, models from Deepseek-r1 family do not exhibit competitive performance: even the largest Deepseek-r1:32B model achieves only 85.17\% query accuracy, which is inferior to that of the Gemma3:12B (87.56\%) and Qwen3:4B (86.21\%) models.

\begin{table*}[htpb]
	\centering
	\caption{Experimental results of CoP-QA-S using Gemma3 family.}
	\label{tab-cop-s-gemma}
	\begin{tabular}{lccccccccr}
		\hline\hline
		\textbf{Talk2DM using} & \multicolumn{2}{c}{Gemma3:1B} & \multicolumn{2}{c}{Gemma3:4B} & \multicolumn{2}{c}{Gemma3:12B} & \multicolumn{2}{c}{Gemma3:27B} & \multirow{2}{*}{$\mathcal{B}_{\mathcal{T}_i}$}\\ \cline{2-9} 
		\textbf{Query task}    & $\mathcal{A}_C$            & $\mathcal{A}_Q$            & $\mathcal{A}_C$            & $\mathcal{A}_Q$            & $\mathcal{A}_C$             & $\mathcal{A}_Q$            & $\mathcal{A}_C$             & $\mathcal{A}_Q$            \\ \hline
		(1) velocity               & 11.56         & 1.62          & 95.13         & 76.06         & 99.19          & 92.29         & 96.75          & 94.12   & 0     \\
		(2) acceleration           & 0             & \uline{0}             & 20.93         & \uline{19.32}         & 89.54          & \uline{80.89}         & 93.56          & 91.95     & -2   \\
		(3) heading                & 0             & \uline{0}             & 81.19         & 39.60         & 98.42          & 91.88         & 98.22          & 92.67     & -1    \\
		(4) color                  & 95.78         & 32.28         & 100           & 76.79         & 100            & 91.35         & 100            & \uwave{100}      & +1     \\
		(5) classification         & 0             & \uline{0}             & 100           & 80.65         & 100            & 89.41         & 100            & 84.11    & -1    \\
		(6) size                   & 51.05         & 38.91         & 100           & 73.64         & 100            & \uwave{93.10}         & 100            & 93.72     & +1    \\
		(7) status                 & 0             & \uline{0}             & 98.30         & 61.06         & 89.57          & 71.91         & 90.00          & \uline{83.83}      & -2  \\
		(8) distance               & 0             & \uline{0}             & 85.81         & 69.40         & 98.89          & 87.80         & 100            & 98.23     & -1   \\
		(9) count                  & 89.90         & 63.92         & 99.18         & 79.18         & 98.76          & 88.45         & 99.18          & 96.49     & 0   \\
		(10) existence              & 99.85         & 76.56         & 100           & \uwave{90.10}         & 93.01          & 88.50         & 98.25          & 95.78     & +1   \\ \hline
		\textbf{Average}       & 34.81         & 21.33         & 88.05         & 66.58         & 96.74          & 87.56         & 97.60          & 93.09   & --     \\ \hline\hline
	\end{tabular}
\end{table*}

\begin{table*}[htpb]
	\centering
	\caption{Experimental results of CoP-QA-S using Deepseek-r1 family.}
	\label{tab-cop-s-deepseek}
	\begin{tabular}{lccccccccr}
		\hline\hline
		\textbf{Talk2DM using} & \multicolumn{2}{c}{Deepseek-r1:1.5B} & \multicolumn{2}{c}{Deepseek-r1:8B} & \multicolumn{2}{c}{Deepseek-r1:14B} & \multicolumn{2}{c}{Deepseek-r1:32B} & \multirow{2}{*}{$\mathcal{B}_{\mathcal{T}_i}$} \\ \cline{2-9} 
		\textbf{Query task}    & $\mathcal{A}_C$                & $\mathcal{A}_Q$               & $\mathcal{A}_C$               & $\mathcal{A}_Q$              & $\mathcal{A}_C$               & $\mathcal{A}_Q$               & $\mathcal{A}_C$               & $\mathcal{A}_Q$               \\ \hline
		(1) velocity               & 63.69             & 24.02            & 90.38            & \uwave{75.00}           & 65.52            & 63.29            & 78.92            & \uline{77.45}       & -1    \\
		(2) acceleration           & 9.04              & \uline{1.20}             & 58.62            & 53.79           & 60.16            & \uline{56.94}            & 82.83            & 81.31      & -2      \\
		(3) heading                & 19.10             & 5.62             & 92.99            & \uline{50.32}           & 72.28            & 65.15            & 92.20            & 85.37    & -1        \\
		(4) color                  & 75.00             & 20.00            & 87.77            & 73.39           & 80.38            & 79.11            & 90.16            & 89.62       & 0   \\
		(5) classification         & 70.91             & 27.88            & 85.21            & 71.13           & 76.99            & 73.93            & 87.88            & 87.88      &  0     \\
		(6) size                   & 70.30             & 22.42            & 84.17            & 71.22           & 77.82            & 74.69            & 92.89            & \uwave{91.88}     &  +1     \\
		(7) status                 & 55.03             & 16.57            & 77.63            & 61.18           & 63.62            & 58.94            & 69.70            & 68.18       &  0    \\
		(8) distance               & 59.48             & 16.99            & 88.89            & 74.07           & 69.40            & 66.52            & 91.71            & 90.61     & 0      \\
		(9) count                  & 70.66             & 34.13            & 90.48            & 82.31           & 73.61            & 69.90            & 88.89            & 87.88     & 0      \\
		(10) existence              & 76.54             & \uwave{38.68}            & 49.30            & 45.07           & 90.39            & \uwave{88.21}            & 93.29            & 91.52      & +2     \\ \hline
		\textbf{Average}       & 56.97             & 20.75            & 80.54            & 65.75           & 73.02            & 69.67            & 86.85            & 85.17     &  --     \\ \hline\hline
	\end{tabular}
\end{table*}

\subsubsection{Efficiency}
\begin{figure}[htpb]
	\centering
	\includegraphics[scale=0.7]{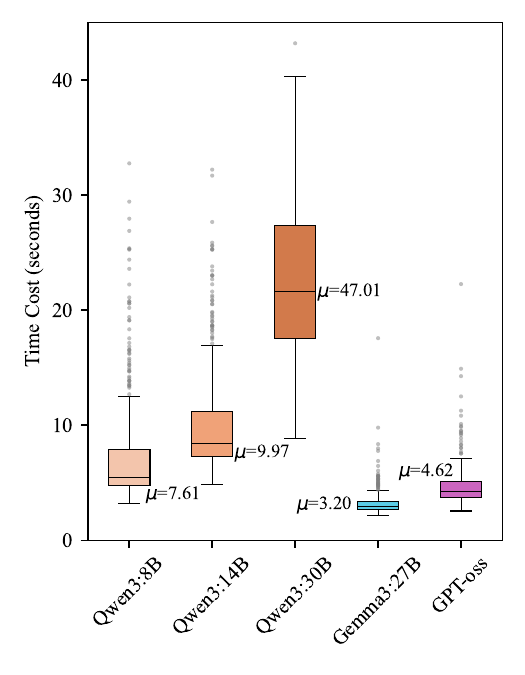}
	\caption{Query time cost of Talk2DM with selected LLMs.}
	\label{fig-timecost}
\end{figure}

\cref{fig-cop-qa-s} shows that, to achieve a query accuracy exceeding 90\%, the Qwen3:8B, Qwen3:14B, Qwen3:30B, Gemma3:27B and GPT-oss models are potentially applicable to Talk2DM. This experiment evaluates the time cost of Talk2DM's NLS querying function when deployed with these models. As shown in \cref{fig-timecost}, Gemma3:27B is the fastest, with a mean query time of $\mu = 3.2$ s, followed by GPT-oss ($\mu=4.62$ s), and Qwen3:8B ($\mu=7.61$ s). These three models are considered suitable for Talk2DM, as they achieve a favorable trade-off between query accuracy and efficiency, making them potentially deployable on edge devices. In contrast, although Qwen3:14B and Qwen3:30B provide only marginal accuracy gains, they incur substantial efficiency degradation and are therefore deemed unsuitable.

Finally, we summarize the most important experimental conclusions in \cref{tab-best}, which shows that GPT-oss and Gemma3:27B are the best underlying LLMs for Talk2DM, completing accurate NLS querying within 2$\sim$5 s.

\begin{table}[htpb]
	\centering
	\caption{The most suitable configurations of Talk2DM.}
	\label{tab-best}
	\begin{tabular}{cccc}
		\hline\hline
		\textbf{Talk2DM using}                           & \textbf{Qwen3:8B} & \textbf{Gemma3:27B} & \textbf{GPT-oss} \\ \hline
		\multicolumn{1}{c|}{\textbf{Accuracy} $\uparrow$}   & 94.34\%    & 93.09\%      & 94.45\%   \\
		\multicolumn{1}{c|}{\textbf{Efficiency} $\downarrow$} & 7.61s     & 3.20s       & 4.62s    \\ \hline\hline
	\end{tabular}
\end{table}

\section{Conclusion}
In this paper, we propose VRCsim, a VRC-DM simulation framework that generates virtual onboard and roadside sensor data streams and aggregates them in the cloud. Based on VRCsim, we construct VRC-QA, a dataset comprising 10K VRC-CP scenes and 100K QA pairs on spatial queries, to facilitate research on VRC-CP question answering. Building upon VRCsim and VRC-QA, we further introduce Talk2DM, a human interface that enables OTH and NLS querying, as well as commonsense reasoning in VRC-DM systems. Talk2DM overcomes the NLoS and FoV limitations of conventional standalone ADS, while enhancing the flexibility and diversity of information retrieval through DMs. Within Talk2DM, we propose a CoP mechanism to guide underlying LLMs in accurately performing NLS querying over structured CP data in VRC-DM. We conduct extensive experiments on VRC-QA to evaluate Talk2DM, and the results demonstrate its strong generalization capability, as it seamlessly adapts to different LLMs while consistently achieving query accuracy above 90\%. In contrast, classical one-shot prompting methods attain only about 30\% accuracy. Moreover, the experimental results reveal a clear trade-off: larger models tend to deliver higher query accuracy while incurring substantially greater query latency. Considering both accuracy and efficiency, Gemma3:27B and GPT-oss emerge as the most suitable LLMs for Talk2DM, achieving over 93\% query accuracy while maintaining high querying efficiency in natural language interactions, with average response time of approximately 2$\sim$5 s.



\bibliographystyle{IEEEtran}
\bibliography{IEEEabrv, ref}

\end{document}